\documentclass[letterpaper]{article} 
\usepackage{aaai25}  
\usepackage{times}  
\usepackage{helvet}  
\usepackage{courier}  
\usepackage[hyphens]{url}  
\usepackage{graphicx} 
\usepackage{natbib}  
\usepackage{caption} 
\usepackage{algorithm}
\usepackage{algorithmic}
\usepackage{amsmath}
\usepackage{amsthm,amsmath,amssymb}
\usepackage{mathrsfs}
\usepackage{booktabs}
\usepackage{pifont}       
\usepackage{bbding}       
\usepackage{fontawesome}  
\usepackage{multirow}

\newcommand{\eat}[1]{}
\usepackage{newfloat}
\usepackage{listings}
\DeclareCaptionStyle{ruled}{labelfont=normalfont,labelsep=colon,strut=off} 
\floatstyle{ruled}
\newfloat{listing}{tb}{lst}{}
\floatname{listing}{Listing}
\title{Enhancing Multi-Robot Semantic Navigation Through Multimodal Chain-of-Thought Score Collaboration}
\author{
    Zhixuan Shen,
    Haonan Luo\thanks{Corresponding author},
    Kexun Chen,
    Fengmao Lv,
    Tianrui Li
}
\affiliations{
    School of Computing and Artificial Intelligence, Southwest Jiaotong University, China\\

    \{shenzx29, chenkexun\}@my.swjtu.edu.cn, \{lhn, trli\}@swjtu.edu.cn, fengmaolv@126.com
}

\usepackage{bibentry}

\begin{document}
\maketitle
\begin{abstract}
Understanding how humans cooperatively utilize semantic knowledge to explore unfamiliar environments and decide on navigation directions is critical for house service multi-robot systems. Previous methods primarily focused on single-robot centralized planning strategies, which severely limited exploration efficiency. Recent research has considered decentralized planning strategies for multiple robots, assigning separate planning models to each robot, but these approaches often overlook communication costs. In this work, we propose Multimodal Chain-of-Thought Co-Navigation (MCoCoNav), a modular approach that utilizes multimodal Chain-of-Thought to plan collaborative semantic navigation for multiple robots. MCoCoNav combines visual perception with Vision Language Models (VLMs) to evaluate exploration value through probabilistic scoring, thus reducing time costs and achieving stable outputs. Additionally, a global semantic map is used as a communication bridge, minimizing communication overhead while integrating observational results. Guided by scores that reflect exploration trends, robots utilize this map to assess whether to explore new frontier points or revisit history nodes. Experiments on HM3D\_v0.2 and MP3D demonstrate the effectiveness of our approach. Our code is available at https://github.com/FrankZxShen/MCoCoNav.git.
\end{abstract}

%

\begin{figure}[t]
    \centering
    \includegraphics[width=8.5cm]{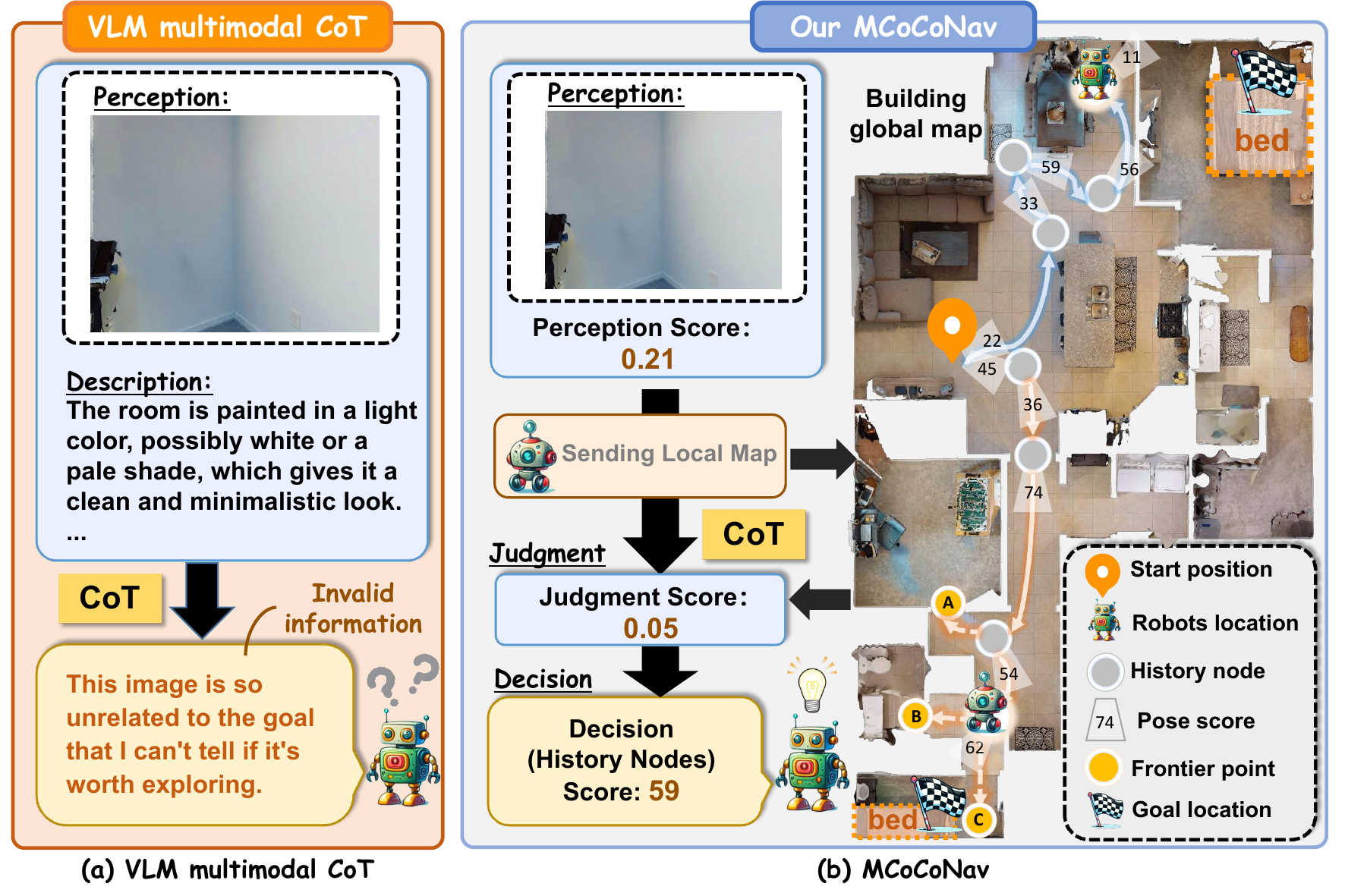}
    \caption{Examples of (a) VLM multimodal CoT reasoning and (b) MCoCoNav cross-image multimodal CoT reasoning. Our MCoCoNav utilizes cross-image multimodal CoT to facilitate robots' simultaneous understanding of scene perspectives and the global semantic map for effective zero-shot multi-robot semantic navigation.}
    \label{fig:intro-1}
\end{figure}

\section{Introduction}
The capability to navigate to designated goals is crucial for house service robots, enabling them to effectively find designated objects in unfamiliar indoor environments and complete various subsequent tasks. Consequently, the Object Goal Navigation (ObjectNav)~\cite{du2020learning,mayo2021visual,chaplot2020object} task has garnered significant attention. Traditional ObjectNav tasks require robots to be navigated to user-specified categories of objects in an unseen and unmapped environment based on visual observations. Given that the environment is invisible to all robots, they must collaboratively infer potential positions where the goals may appear. This necessitates effective communication and cooperation among multiple robots (e.g., conflict-free communication and global planning after communication), enabling them to make corresponding decisions based on the observed visual cues.

To establish a structured navigation cooperation framework for multiple robots, centralized planning strategies~\cite{zhao2024hierarchical,agashe2023evaluating,yu2023co,Chen2023Scalable} map all robots' observations, histories, and other pertinent information to a unified channel, with a single planning model tasked with assigning objectives to each robot group or individual. However, as the complexity of the environment and the number of robots increases, the information processing burden on the planning model significantly escalates. Alternatively, decentralized planning  strategies~\cite{Chen2023Scalable,ying2024goma,liu2023llm,wang2024safe} assign each robot an individual "brain" for independent reasoning, allowing the robots to communicate and share information about explored areas similarly to how humans do. These approaches facilitate the connection between new discoveries and previously explored regions, enabling adaptive decisions. By distributing the information generated by multiple robots across various planning models, the burden of decision is alleviated. Nevertheless, the decentralized planning strategy is still limited due to the substantial communication and temporal costs associated.

Utilizing VLMs as tools for multimodal scene understanding and navigation planning provides easily interpretable intermediate representations for robotic systems. Although the narratives generated by VLMs may not be sufficient for navigation, their Chain-of-Thought (CoT) reasoning methods, which simulate human thought processes, can inform or guide the behavior of the underlying navigation stack in multi-robot semantic navigation tasks (e.g., providing intermediate representations for information processing and communication among robots). Consequently, some work~\cite{kuang2024openfmnav,shah2023navigation,ren2024explore} has integrated VLM multimodal CoT~\cite{wei2022chain,zhang2023multimodal,zheng2023ddcot,gao2024cantor} into centralized planning strategies, using the problem decomposition reasoning of multimodal CoT as a heuristic to specify strategies. However, utilizing multimodal CoT for navigation in complex and diverse indoor scenes may be unreliable. As shown in Figure \ref{fig:intro-1} (a), meaningless scene perspectives frequently disrupt the multimodal CoT reasoning process during navigation. In contrast, if both the scene perspectives and the robot's global semantic map in a decentralized planning strategy can be comprehended by multimodal CoT, the reasoning decisions will be more reliable. For instance, the location of the scene perspective on the global semantic map can be used to infer the appropriate answer.

In order to make high-level information from scene perspectives and global semantic map available for multimodal CoT to understand, we propose Multimodal Chain-of-Thought Co-Navigation (MCoCoNav), a novel framework that utilizes multimodal Chain-of-Thought to develop effective exploration and decision strategies for multi-robot navigation in unfamiliar environments. As shown in Figure \ref{fig:intro-1} (b), given the current scene perspective of a robot, the Perception module of MCoCoNav utilizes multimodal CoT to evaluate its exploration value, predicting the probability of "Yes" as an exploration score. Considering the significant communication costs incurred during navigation planning, MCoCoNav uses the global semantic map as the direct communication bridge between robots, avoiding additional communication cost overhead. All robots jointly maintain a global semantic map that integrates all observations of the unseen environment. Furthermore, to fully leverage the role of history nodes in the global semantic map, for each node being explored by a robot, we examine the VLM's inclination to explore frontier points and design metrics to calculate the horizontal field of view score and history score of the node. The former indicates the robot's tendency to explore the current node, and the latter indicates the exploration possibilities of different history nodes, both of which are synchronized with the update of the global semantic map. Subsequently, each robot selects the frontier point with the highest prediction probability of the Decision module or the history node with the highest history score as its long-term navigation goal on the global semantic map. 

Extensive experiments on benchmark navigation metrics and evaluations on HM3D\_v0.2 and MP3D demonstrate that MCoCoNav efficiently plans collaborative exploration of multiple robots in unfamiliar indoor environments, outperforming other multi-robot methods based on centralized and decentralized planning strategies. Our contributions can be summarized as follows: 
(1) We design an innovative planning framework for the multi-robot semantic navigation task that enables local, small-scale VLMs to guide multiple robots through unknown environments for efficient exploration and decision.
(2) Our proposed cross-image multimodal CoT facilitates robots' understanding of high-level information from different images and enables low-cost semantic information sharing among robots.
(3) Evaluations on HM3D\_v0.2 and MP3D demonstrate the superior performance of MCoCoNav, which is fully zero-shot and can be deployed locally at low cost.

\section{Related Work}
\noindent \textbf{Zero-Shot Object Goal Navigation} Finding specified objects in unfamiliar indoor environments is a widely applied task for embodied intelligent robots. While numerous works have utilized reinforcement learning~\cite{ye2021auxiliary,chang2020semantic}, learning from demonstrations~\cite{ramrakhya2023pirlnav}, or waypoint planners~\cite{ChenLKGY23,chaplot2020object,zhang2021hierarchical,luo2022stubborn,ramakrishnan2022poni,zhang20233d} to train robots with semantic navigation capabilities, these task-specific training methods are difficult to generalize to diverse real-world environments. Consequently, many researchers have begun to investigate Zero-Shot Object Navigation (ZSON) techniques. ZSON~\cite{majumdar2022zson} and COWs~\cite{gadre2023cows} employ CLIP~\cite{radford2021learning} features or open-vocabulary object detectors to locate goal objects. Recent studies such as LGX~\cite{dorbala2023can}, ESC~\cite{zhou2023esc}, L3mvn~\cite{yu2023l3mvn} and Voronav~\cite{wu2024voronav} involve using Large Language Models (LLMs) for reasoning and decision. However, these works only consider a single robot decision framework and rely on powerful remote large foundation models like GPT-4V~\cite{yang2023dawn}. In contrast, we use only a quantized 9B lightweight VLM to formulate multi-robot collaborative decision strategies.

\noindent \textbf{LLM-based Multi-Robot Systems} Compared to more mature approaches such as active mapping~\cite{ye2022multi}, reinforcement learning planning~\cite{yu2022learning}, and methods based on prior knowledge~\cite{liu2022multi}, the application of LLMs in Multi-Robot Systems (MRS) is still in its early stages. Nevertheless, a few contemporary studies have begun to explore the use of generative models in MRS task planning. Some research adopts the centralized planning strategy, where the LLM needs to simultaneously comprehend the observations, histories, and task states of multiple robots and allocate collaborative tasks to each robot. Specifically, Co-NavGPT~\cite{yu2023co} utilizes a single LLM to assign exploration frontiers to each robot. HAS~\cite{zhao2024hierarchical} automatically organizes LLM-based robot groups to complete navigation tasks in complex Minecraft environments. Furthermore, other studies employ the decentralized planning strategy, treating each robot as a separate entity that exchanges information through methods similar to human thinking and communication and makes independent decisions. For instance, CoELA~\cite{zhang2023building} provides a systematic template for decentralized communication and collaboration, while GOMA~\cite{ying2024goma} frames the language interaction between robots and humans as a planning problem.

\noindent \textbf{Multimodal CoT Reasoning}  Effective prompting techniques are crucial for fully harnessing the capabilities of LLMs and VLMs. Recent advancements~\cite{diao2023active,ho2022large,wang2022self,zhang2022automatic} have introduced CoT to further enhance the reasoning abilities of LLMs. Concurrently, researcher~\cite{zhang2023multimodal,zheng2023ddcot,gao2024cantor}s have focused on maximizing the multimodal reasoning potential of LLMs and VLMs through multimodal CoT. Our work transcends the exploration of simple CoT reasoning strategies, focusing instead on information exchange and collaborative navigation among multiple robots using multimodal CoT.

\begin{figure*}[!t]
\centering
\includegraphics[width=1.0\textwidth]{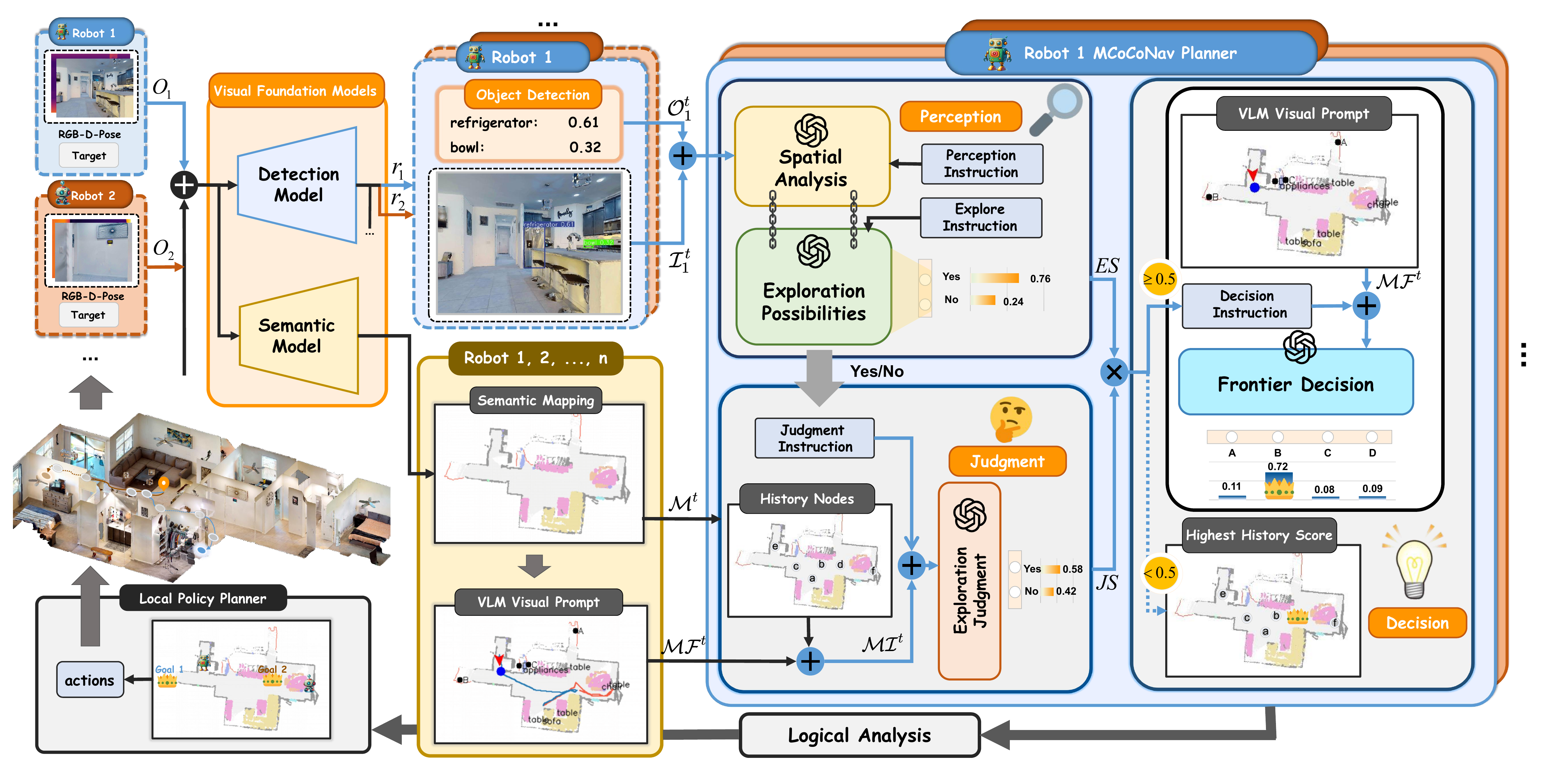}
	\caption{\textbf{Components of MCoCoNav.} The MCoCoNav architecture consists of Visual Foundation Models, the MCoCoNav Planner and the Local Policy Planner. At its core is the MCoCoNav Planner, which is composed of three main components: the Perception module, the Judgment module and the Decision module.
 }
\label{fig:method-1}
\end{figure*}

\section{Method}
\subsection{Problem Formulation}
\subsubsection{Multi-Robot Semantic Navigation} In the Multi-Robot Semantic Navigation task, the set of scenes can be denoted as $S = \left\{s_{1}, ..., s_{k}\right\}$, and the set of categories as $C = \left\{c_{1}, ..., c_{m}\right\}$. In each episode, $n$ robots $R = \left\{r_{1}, ..., r_{n}\right\}$ are randomly placed in an unfamiliar and invisible scene $s_{i}$, with the objective of locating an object of category $c_{i}$. Each robot is allowed a maximum of 500 time steps per episode. At each time step $t$, robot $r_{i}$ obtains a current observation $O$, which comprises an RGB-D image, the robot's location, and pose. The action space $A$ consists of six discrete actions: \texttt{move\_forward}, \texttt{turn\_left}, \texttt{turn\_right}, \texttt{look\_up}, \texttt{look\_down}, and \texttt{stop}. When executing the \texttt{move\_forward} action, the robot moves 25$cm$ ahead, while for \texttt{turn\_left}, \texttt{turn\_right}, \texttt{look\_up}, or \texttt{look\_down}, the robot rotates 30° in the corresponding direction. The robot must select and execute an action from the action space $A$. The stop action is triggered when one of the robots $r_{i}$ approaches the goal object. An episode is considered successful if the distance between robot $r_{i}$ and the goal is less than 0.2m and robot $r_{i}$ executes the stop action. In this task robots are not allowed to use purposefully trained models and panoramic 360° FoV sensors.
\subsubsection{VLM Predictions} VLMs that have undergone large-scale pretraining provide the necessary reasoning capabilities to address the multi-robot semantic navigation task. Given an RGB image $I$ and text $T$, we query the VLM to predict the probability of the next token $y$. Specifically, we denote the VLM's prediction as $\hat{f}_{y}(I,T) \in [0, 1]^{|Y|}$, which represents the softmax scores over a selection set $Y$.

\subsection{Overview}
Our MCoCoNav methodology is illustrated in Figure \ref{fig:method-1}. The core of the MCoCoNav is the MCoCoNav Planner, which consists of three main components. We begin by employing the Perception module to analyze whether each robot's current perspective warrants exploration, assigning the exploration score. Concurrently, the Judgment module evaluates whether the robot should explore predicted frontier points or return to history nodes for re-exploration based on the global semantic map of all robots, history nodes, predicted frontier points and trajectory information, assigning the judgment score. Subsequently, we combine the exploration and judgment scores, which are weighted to produce the horizontal field of view score and history score for the current history node. Suppose the horizontal field of view score exceeds or is equal to 0.5. In that case, the Decision module integrates information from the Perception and Judgment module to select an appropriate frontier point as the long-term navigation goal for the next time step. If the horizontal field of view score is below 0.5, the robot's long-term navigation goal for the next time step is set to the history node with the highest history score. With long-term navigation goals established for each robot, we logically analyze the navigational feasibility of these goals and employ the local policy to guide each robot in exploration and goal acquisition.

\subsection{MCoCoNav Planner}
The MCoCoNav Planner is a \textbf{Decentralized Planning} system that interacts with a single scene view, a single navigation goal text, a global semantic map, and global navigation history nodes to perform long-term goal prediction for individual robot navigation. In this section, we will introduce the planning process of the MCoCoNav Planner.

\subsubsection{Preliminaries}
For robot $r_i$ at time step $t$, the obtained observation $O$ includes the scene view $I^{t}_{i}$, scene semantic map $S^{t}_{i}$, scene depth map $D^{t}_{i}$ and pose $p$. We first process the visual information of individual robots using Visual Foundation Models (VFMs) to obtain object detection information $\mathcal{O}^{t}_{i}$ and semantic maps $\mathcal{M}^{t}$ for MCoCoNav Planner's Visual Perception and Global Map Exploration Judgment. We provide details of object detection and semantic mapping in Appendix A.1. 

\subsubsection{Visual Perception} \label{visual}
We guide the VLM's prediction through a simple multimodal CoT to obtain the exploration score ($ES$) for the current perspective. Firstly, we use the Perception Instruction $\mathcal{I}_{P}$, a query template containing the query \textit{Object List}, \textit{Spatial Relationships} and \textit{Additional Context}, to prompt the Perception VLM to provide a description of the spatial relationships in the scene view $I^{t}_{i}$. 
Specifically, we query the Perception VLM: \textsl{"Are they to your left, right, in front of you, or behind you?"} and \textsl{"Are they on the floor, mounted on the wall, or placed on top of another piece of furniture?"}, and \textsl{"How close are they to you or to each other?"}.
Subsequently, we combine VLM output text $VLM(I^{t}_{i},\mathcal{I}_{P})$ with the object detection information $\mathcal{O}^{t}_{i}$ in pure natural language to form the multimodal CoT to infer whether the current scene is worth further exploration by using the Exploration Instruction $\mathcal{I}_{E}$ containing \textit{Target of Navigation}, \textit{Scene Objects} and \textit{Decision Criteria} information. It is noteworthy that we only extract the normalized probability output in the "Yes" direction from this CoT to construct the $ES^{t}_{i,p}$ for the current pose $p$:
\begin{align}
    ES^{t}_{i,p} &= \hat{f}_{"Yes"}(I^{t}_{i},VLM(I^{t}_{i},\mathcal{I}_{P})+\mathcal{O}^{t}_{i}+\mathcal{I}_{E})
\end{align}

\subsubsection{Global Map Exploration Judgment} \label{judgment}
Now, we will detail the Judgment module. If the global map $\mathcal{M}^t$ has just been initialized, all robots are set to continue exploring. Otherwise, prompts need to be constructed for Judgment VLM to determine whether it is worthwhile for the robots to continue exploring.

To obtain the visual prompt, first, we annotate the current position coordinates and pose of robot $r_i$, as well as its last long-term navigation goal, with a red arrow and a blue dot respectively. Simultaneously, we sample the coordinates of all robots' history nodes with lowercase letters in chronological order (in the rare event that the number of history nodes exceeds 26, we use lowercase letters followed by the sequence number minus 26 for those beyond 26). Then, we annotate the predicted frontier points with uppercase letters in the same manner. Finally, we map all annotation information onto the global semantic map $\mathcal{M}^t$, resulting in the visual prompt $\mathcal{MI}^{t}$ that contains semantic and historical information for all robots.

To obtain the textual prompt, we construct the Judgment Instruction $\mathcal{I}_{J}$ by incorporating the input-output chain from the Perception module with a query template that includes information on \textit{Frontier Points}, \textit{History Nodes} and \textit{Location and Previous Movement}. Unlike the Perception module, the "yes" output of the VLM corresponds to "explore a frontier point". The judgment score ($JS^{t}_{i,p}$) of the current pose $p$ can be represented as:
\begin{align}
 JS^{t}_{i,p} &= \hat{f}_{"Yes"}(\mathcal{MI}^{t}, \mathcal{I}_{J})
\end{align}

We apply temperature scaling ($\tau_{ES}$ and $\tau_{JS}$) for judgment score and exploration score and compute the current horizontal field of view score $HFOVS^{t}_{i,hfov}$:
\begin{align}
    HFOVS^{t}_{i,hfov} &= exp(\tau_{ES} \cdot ES^{t}_{i,p} + \tau_{JS} \cdot JS^{t}_{i,p})
\end{align}
where $hfov$ represents the current horizontal field of view. If $HFOVS^{t}_{i,hfov}$ is greater than or equal to 0.5, robot $r_i$ continues to select frontier points for exploration. Otherwise, robot $r_i$ returns to history nodes for re-exploration.

\begin{figure*}[!t]
\centering
\includegraphics[width=1.0\textwidth]{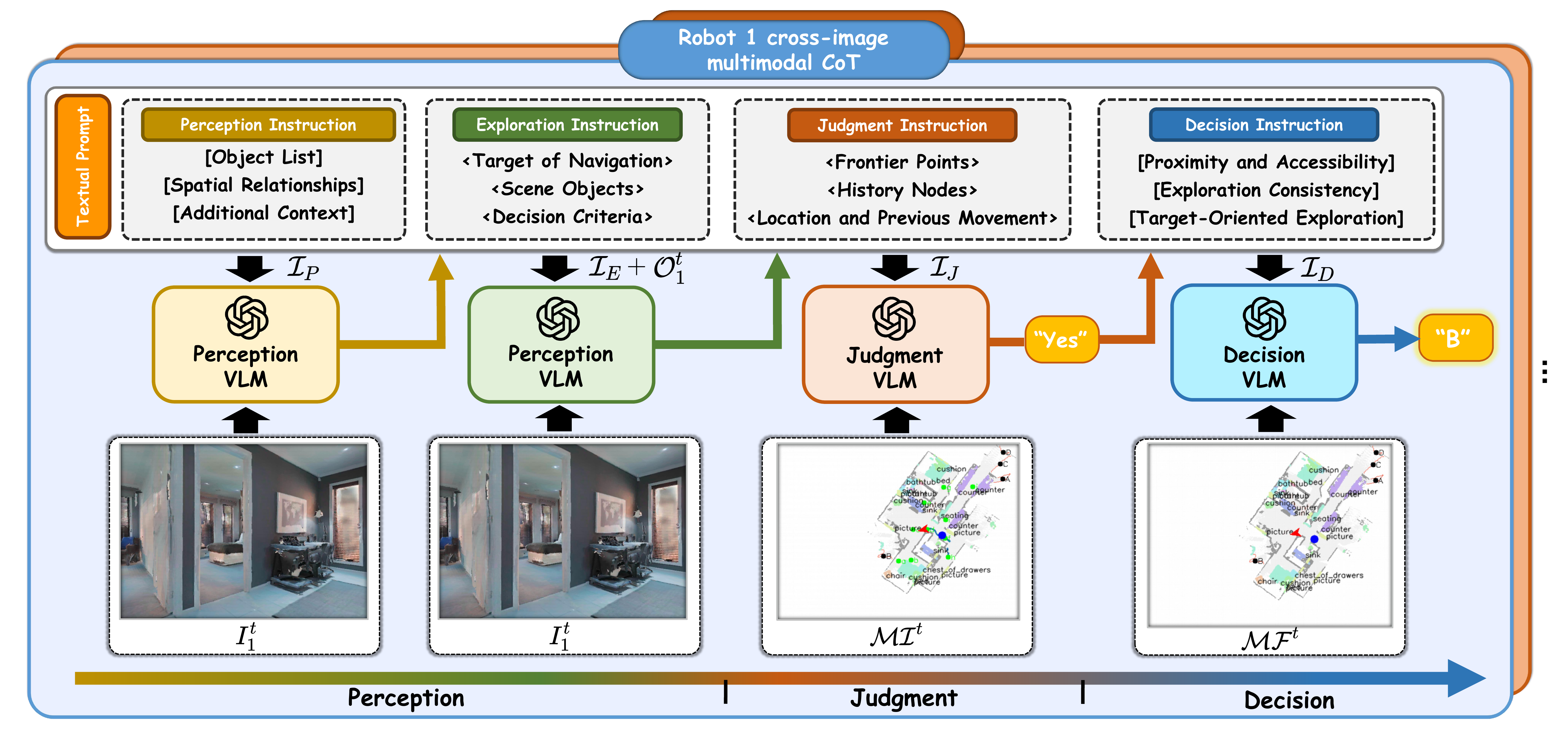}
	\caption{
	\textbf{Workflow of cross-image multimodal CoT.} Information from multiple images is unified into a multimodal CoT through VLM. At each time step, the next action for each robot and the global semantic map are updated based on the observed scene objects. The cross-image multimodal CoT enables semantic label alignment for navigation planning, environmental understanding, and common-sense reasoning.}
\label{fig:method-2}
\end{figure*}

\begin{table*}
\centering
\begin{tabular}{lcccccc}
\toprule
\multirow{2}{*}{Method} & \multirow{2}{*}{\shortstack{Zero-Shot}} & \multirow{2}{*}{\shortstack{LLM/VLM}} & \multicolumn{2}{c}{HM3D\_v0.2} & \multicolumn{2}{c}{MP3D} \\
\cmidrule{4-7}
&  &  & SPL$\uparrow$ & SR$\uparrow$ & SPL$\uparrow$ & SR$\uparrow$ \\
\midrule
Greedy~\cite{visser2013discussion} & \ding{51} & None & 0.328 & 0.611 & 0.225 & 0.406 \\
Cost-Utility~\cite{julia2012comparison} & \ding{51} & None & 0.323 & 0.625 & 0.212 & 0.419 \\
Random Sampling & \ding{51} & None & 0.336 & 0.636 & 0.281 & 0.435 \\
Multi-SemExp~\cite{chaplot2020object} & \ding{55} & None & 0.327 & 0.612 & - & - \\
Co-NavGPT~\cite{yu2023co} & \ding{51} & GPT-3.5 (Remote) & 0.331 & 0.661 & - & - \\
MCoCoNav (Ours) & \ding{51} & GLM-4V (Local) & \textbf{0.387} & \textbf{0.716} & \textbf{0.334} & \textbf{0.568} \\
\midrule
Co-NavGPT (GT-Seg) & \ding{51} & GPT-3.5 (Remote) & 0.448 & 0.757 & - & - \\
MCoCoNav (GT-Seg) & \ding{51} & GLM-4V (Local) & \textbf{0.464} & \textbf{0.872} & \textbf{0.410} & \textbf{0.655} \\
\bottomrule
\end{tabular}
\caption{\textbf{Comparison of different baselines for 2-robot on HM3D\_v0.2 and MP3D.} MCoCoNav significantly outperforms all baseline methods on all metrics and achieves zero-shot multi-robot semantic navigation for local planning.}
\label{tab:table1}
\end{table*}
\subsubsection{Cross-image Multimodal CoT Decision} \label{decision}
We have been discussing the utilization of VLM to judge exploration possibilities. However, we haven't addressed the most crucial step in completing exploration—determining the next long-term navigation goal, which involves selecting appropriate frontier points or history nodes. VLM's evaluation of suitable long-term navigation goals depends on the current position of robot $r_i$ and the history scores ($HS$) of all robots' history nodes.

We get the history score $HS$ in the following steps. First, suppose $r_i$ is at a distance greater than or equal to 25 pixels from the nearest history node. In that case, we construct a 360-dimensional zero vector representing the state score of the robot's current position and fill the $HFOVS^{t}_{i,hfov}$ of robot $r_i$ into this vector corresponding to the robot's angle of horizontal view. We then sum the entire vector and divide by the number of times all robots have explored within a 25-pixel range of the current position, obtaining the $HS$ for the current position. If robot $r_i$ is less than 25 pixels from the nearest history node, the $HFOVS^{t}_{i,hfov}$ is filled into the state score of that history node and the $HS$ is updated. For the readers' convenience, we present the complete history score algorithm in Appendix A.2. After obtaining the $HS$, if the robot $r_i$ is selected to return to the history nodes, we will choose the history node with the highest $HS$ as the long-term navigation goal. Otherwise, we will construct visual and textual prompts for Decision VLM to select the appropriate frontier.

For the visual prompt of Decision VLM, we remove the history nodes annotations from the visual prompt $\mathcal{MI}^{t}$ of Judgment VLM, retaining only the annotations for frontier points. For the textual prompt of Decision VLM, we construct the Decision Instruction $I_D$ by integrating the input-output from the Judgment VLM with a query template that includes considerations of \textit{Proximity and Accessibility}, \textit{Exploration Consistency}, and \textit{Target-Oriented Exploration}. As shown in Figure \ref{fig:method-2}, the continuous multimodal CoT ensures that the VLM can fully comprehend information from multiple images, alleviating the potential invalid information that might arise from a single CoT decision. To derive the final decision result, we limit the VLM's output to four or fewer options representing the frontier points "A", "B", "C" and "D", and then obtain the normalized probabilities for these four directions:
\begin{align}
    DS^{t}_{i} &=    \hat{f}_{X}(\mathcal{MF}^{t}, \mathcal{I}_{D}) 
\end{align}
where $\mathcal{MF}^{t}$ represents the visual prompt at time step $t$ that includes annotations of frontier points, $DS$ signifies the decision score corresponding to the frontier points, and $X$ is the uppercase letter representing the respective frontier points. We select the frontier point represented by the uppercase letter with the highest decision score as the long-term navigation goal. 

\subsection{Logical Analysis and Local Policy} \label{local}
Even though the long-term navigation goal has been determined, the aforementioned process does not consider two logic issues during navigation: (1) The continuity of exploration towards the long-term navigation goal. (2) Potential collisions. 

To address the first issue, each time the MCoCoNav Planner execution ends, we assess whether the robot has reached the previous long-term navigation goal. If the robot is at a distance of 25 pixels or more from the last long-term navigation goal and the $HFOVS^{t}_{i,hfov}$ is less than 0.5, we consider that the robot has not reached the goal and that the current position is not worth exploring. In this case, the robot needs to continue exploring towards the previous long-term navigation goal. For the second issue, we record the relative distance between the current robot position and the position at the end of the last MCoCoNav Planner execution. If this distance is less than 25 pixels, we consider that the robot is trapped in a location, experiencing collisions with nearby objects or obstacles over several time steps. At this point, the long-term navigation goal will be reset to a randomly sampled point on the global semantic map $M^t$.

Finally, we utilize a control strategy based on the Fast Marching Method (FMM)~\cite{sethian1999fast} to output a low-level action $a^{t}_{i}$ for each robot, which concludes the loop and moves to the next time step $t+1$.

\section{Experiment}
\subsection{Experimental Setup}
\noindent \textbf{Datasets.} 
We evaluate our approach using the Habitat~\cite{savva2019habitat} simulator and validate it on two different real-world environment 3D scan datasets: HM3D\_v0.2~\cite{ramakrishnan2021habitat} and MP3D~\cite{chang2017matterport3d}. The validation split for HM3D\_v0.2 includes 1000 episodes, spanning 36 scenes and 6 object categories, while the validation split for MP3D includes 2195 episodes, spanning 11 scenes and 21 object categories.

\noindent \textbf{Metrics.} We employ the navigation success rate (SR) and the success rate weighted by navigation path length (SPL) as evaluation metrics. \textbf{SR} represents the percentage of successful episodes out of the total number of episodes. \textbf{SPL} is calculated as the inverse ratio of the actual path length to the optimal path length weighted by the success rate. 

\noindent \textbf{Implementation Details.} We employ YOLOV10m\cite{wang2024yolov10} as the Detection model and RedNet\cite{jiang2018rednet} as the Semantic model. All VLMs utilize the locally deployed INT4-quantized GLM-4V-9B~\cite{glm2024chatglm}. To learn more about the details of the experiments and the VLM Prompt setup, see Appendix B. 

\subsection{Baselines}
We evaluate MCoCoNav by comparing it with several multi-robot Semantic Navigation baselines: Greedy~\cite{visser2013discussion}, Cost-Utility~\cite{julia2012comparison}, Random Samples, Multi-SemExp~\cite{chaplot2020object}, and Co-NavGPT~\cite{yu2023co}. In the Greedy strategy, each robot selects its nearest designated frontier as its goal location. The Cost-Utility strategy conducts a cost-utility assessment for each frontier cell after obtaining the frontiers, choosing the highest-scoring frontier as the goal location. The Random Samples strategy randomly samples long-term navigation goals on the map. Multi-SemExp is a baseline extended from ~\cite{chaplot2020object} for multi-robot settings. Co-NavGPT employs a centralized planning approach, encoding the explored environmental data as prompts for LLM. 
\begin{figure*}[!t]
\centering
\includegraphics[width=1.0\textwidth]{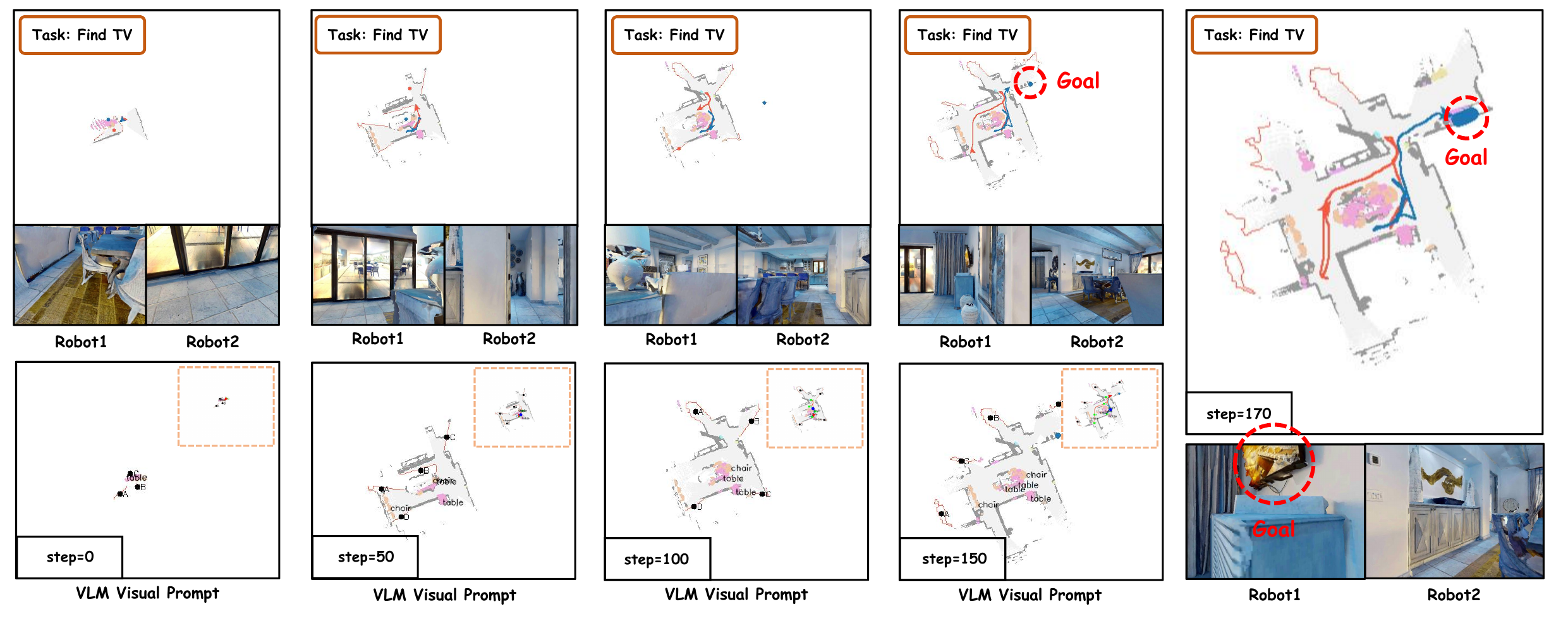}
	\caption{
	\textbf{2-Robot navigation episode with MCoCoNav in HM3D\_v0.2.} The top row shows the robots' RGB scene views and the global semantic map. The bottom row shows the VLM Visual Prompt used for Decision (large image) and the VLM Visual Prompt used for Judgment (small image). Best viewed when zoomed in.
 }
\label{fig:exp-1}
\end{figure*}
\subsection{Results and Analysis}
\subsubsection{Comparison with baseline methods}
The performance of MCoCoNav with 2-robot in comparison to other methods on the HM3D\_v0.2 and MP3D datasets is summarized in Table \ref{tab:table1}. 
Firstly, compared to Multi-SemExp, which is trained end-to-end on local maps, MCoCoNav increases +6.0\% SPL and +10.4\% SR on HM3D\_v0.2. The results demonstrate that our method constructs a global map shared by the multi-robot system, enriching the semantic information of the environment and thereby achieving better performance. 
Compared to other zero-shot methods, such as Co-NavGPT, MCoCoNav achieves +5.6\% SPL and +5.5\% SR on HM3D\_v0.2. Meanwhile, +5.3\% SPL and +13.3\% SR on MP3D compared to Random Sampling. Our MCoCoNav method further improves the navigation process by constructing a comprehensive analysis framework for scene images and global map information, striking a balance between exploration and return, and is better adapted to untrained complex environments and navigation goals.

Furthermore, we evaluate the importance of navigation performance alone without considering the accuracy of semantic segmentation. We replace the semantic segmentation algorithm with ground-truth (GT-Seg) and achieve an increase of +1.6\% SPL and +11.5\% SR on HM3D\_v0.2 compared to Co-NavGPT and +12.9\% SPL and +22.0\% SR on MP3D compared to Random Sampling. This establishes new state-of-the-art metrics for these datasets. Figure \ref{fig:exp-1} illustrates a successful case of MCoCoNav navigating to the goal "TV" using multimodal CoT, visualizing the observations of four key global decisions along with the synchronously updated details of the VLM Visual Prompt and global semantic map.
\begin{table}
\centering
\begin{tabular}{lccccc}
\toprule
Seg. & Robot Num. & T $\downarrow$ & DT$\downarrow$ & SPL$\uparrow$ & SR$\uparrow$ \\
\midrule
Pred-Seg & 1 & 744s & 132 & 0.297 & 0.634 \\
Pred-Seg & 2 & 883s & 156 & 0.387 & 0.716 \\
Pred-Seg & 3 & 1110s & 192 & 0.442 & 0.720 \\
\midrule
GT-Seg & 1 & 342s & 0 & 0.298 & 0.736 \\
GT-Seg & 2 & 582s & 0 & 0.464 & 0.872 \\
GT-Seg & 3 & 961s & 0 & 0.485 & 0.911 \\
\bottomrule
\end{tabular}
\caption{\textbf{Effect of the number of robots and semantic accuracy on HM3D\_v0.2.} 
Pred-Seg represents the semantic segmentation predicted by the model and DT denotes the number of the Detection Trap. T is the average time spent per episode.}
\label{table2}
\end{table}
\subsubsection{Effect of the number of robots and semantic accuracy}
We investigate the impact of varying robot numbers and semantic accuracy on navigation performance on HM3D\_v0.2. As shown in Table \ref{table2}, significant differences are observed under different semantic segmentation conditions. We attribute the limitations of VLM in addressing the multi-robot semantic navigation task to two primary factors: the accuracy of semantic segmentation and the quality of 3D scanning. From the results, we observed that both SR and SPL increase with the number of robots, although the rate of improvement gradually decreases. However, the navigation time cost significantly increases. The frequency of detection traps (DT), where robots incorrectly identify the wrong objects as goals, also rises with an increasing number of robots. Notably, when there are two robots, the improvements in SR and SPL are the greatest, while the increase in detection traps and navigation time is minimal. In addition, since the 3D scan quality of the MP3D is significantly lower than that of the HM3D\_v0.2, its performance in Table \ref{tab:table1} is lower than that of the HM3D\_v0.2.
\begin{table}
\centering
\begin{tabular}{cccc ccc}
\toprule
\multicolumn{5}{c}{\textbf{Modules}} &
\multirow{2}{*}{SPL$\uparrow$} & 
\multirow{2}{*}{SR$\uparrow$} \\
\cmidrule(lr){1-5} 
Per & PCoT & Jud & Dec & Log \\
\midrule
 & & & \ding{51} & & 0.332 & 0.676 \\
 \ding{51} & & & \ding{51} & & 0.349 & 0.677  \\
\ding{51} & \ding{51} & & \ding{51} & & 0.353 & 0.698  \\
& & \ding{51} & \ding{51} & & 0.341 & 0.692 \\
\ding{51} & \ding{51} & \ding{51} & \ding{51} & & 0.381 & 0.707 \\
& & & & \ding{51} & 0.286 & 0.648 \\
\ding{51} & \ding{51} & \ding{51} & \ding{51} & \ding{51} & \textbf{0.387} & \textbf{0.716} \\
\bottomrule
\end{tabular}
\caption{\textbf{Ablation Study.} 
Per means the Perception module without multimodal CoT and PCoT denotes the Perception module's multimodal CoT. Jug represents the Judgment module; Dec is the Decision module. Log denotes Logic Analysis.}
\label{table3}
\end{table}
\subsubsection{Ablation Study}
We present the impact of the three modules and Logical Analysis on MCoCoNav's performance in Table \ref{table3}. The third row of Table \ref{table3} shows that the multimodal CoT of the Perception module is effective, which can alleviate the invalid information that VLM's prediction may cause. Additionally, the Logical Analysis module plays a crucial role in resolving navigation logic problems such as collisions, and when it works together with the other three modules, the performance is significantly improved. The ablation studies demonstrate the effectiveness of each component in our method.

\section{Conclusion}
We propose MCoCoNav, a novel framework leveraging multimodal Chain-of-Thought for efficient exploration and decision in multi-robot semantic navigation. To reduce communication overhead, MCoCoNav utilizes a global semantic map for direct communication among robots. Robots maintain the shared map, integrating observations and using node scores to guide exploration decisions. Experimental results demonstrate MCoCoNav's superior performance compared to previous multi-robot methods.

\bibliography{aaai25}

\newpage

\section{Technical Appendix for MCoCoNav
}
In this appendix, we provide additional details and analysis of our approach. We explain more about the construction of semantic maps and the History Score algorithm in Section A. In addition, we present more details of the experiment in Section B. The template of prompts used for the experiments is provided in Section B.7.
\section{A Method Details}
\subsection{A.1 Object Detection and Semantic Mapping}
The Object Detection information $\mathcal{O}^{t}_{i}$ and Semantic Map $\mathcal{M}^{t}$ These are defined as follows:
\begin{align}
    \mathcal{O}^{t}_{i} &= det(I^{t}_{i}) \\
    \mathcal{M}^{t} &= \sum^{n}_{i=1} map(I^{t}_{i},S^{t}_{i},D^{t}_{i},p)
\end{align}

\noindent \textbf{Object Detection $det$.}
While Vision Language Models (VLMs) possess certain visual perception and expression capabilities, they are unable to provide detailed visual object information  (e.g., predicted probabilities of visual object labels). Therefore, a detection model is required to transform the detailed visual object information of the current environment into natural language descriptions. We employ a detection model as a translator for visual object details, converting visual object labels and predicted probabilities into pure natural language for VLM understanding.

\noindent \textbf{Semantic Mapping $map$.}
At each time step, for robot $r_i$, we project the semantic information obtained from its exploration onto a 2D top-down semantic map. After obtaining $N$ semantic maps from $N$ robots, we normalize these maps to a common coordinate system. The integrated global semantic map $\mathcal{M}^t$ will contain semantic information explored by all robots. 

At each time step, for an individual robot $r_i$, we employ a pre-trained RedNet ~\cite{jiang2018rednet} model to perform semantic segmentation on the given RGB scene view, resulting in a set of $K$ semantic categories. Subsequently, based on the depth image and the pose of robot $r_i$, these $K$ classifications are mapped onto a 3D semantic point cloud and then projected onto a 2D top-down map. All point clouds close to the ground are assigned to an exploration map representing traversable areas, while point clouds at other heights are mapped to an obstacle map. This process ultimately yields a semantic map $m^{t}_{i}$ with dimensions of $(K + 4) \times M \times M$, where $M$ represents the width and height dimensions of the map, and $(K + 4)$ denotes the total number of channels within the map. These channels encompass $K$ classification mappings, an obstacle mapping, an exploration mapping, and mappings of the robot's current and past positions.

At the beginning of each episode, the entire map is initialized to zero, with the robot positioned at the center of the map. After obtaining $N$ semantic maps from $N$ robots, we normalize these maps to a common coordinate system and use the maximum value of each pixel across the individual maps as the pixel value in the merged global map. The integrated global semantic map $\mathcal{M}^t$ will thus contain semantic information from all areas explored by the robots.
\subsection{A.2 History Score Algorithm}
Algorithm \ref{alg1} details the complete computation process for the historical score ($HS$). At each time step, each robot searches for the nearest historical node in its surrounding environment (lines 2-6). If the distance from the robot to the nearest historical node is greater than or equal to 25 pixels, we first check if a state score vector $A$ for the current position already exists. If the state score vector does not exist, we reconstruct a 360-dimensional zero vector $A$ (lines 9-11).  

Next, we fill in the robot $r_i$'s Horizontal Field of View Score $HFOVS^{t}_{i,hfov}$ into the vector corresponding to the robot's angle of horizontal view (lines 12-20). We then sum the entire vector and divide it by the number of exploration attempts made by all robots within 25 pixels of the current position to obtain the $HS$ for the current position (line 21).

If the distance from robot $r_i$ to the nearest historical node is less than 25 pixels, we fill the $HFOVS^{t}_{i,hfov}$ into that historical node and update the $HS$ (lines 22-24).  This approach ensures that the historical score accurately reflects the robot's exploration history and proximity to previously visited areas, thereby aiding in more informed navigation decisions.
\begin{algorithm}[ht]
\caption{History Score Algorithm}
\label{algorithm}
\textbf{Input}: Horizontal
Field of View Score $HFOVS^{t}_{i,hfov}$ of robot $r_i$\\
\textbf{Parameter}: robot $r_i$, current angle of robot $hfov$, current node coordinates $x$, history node set $H$, Euclidean function $D$, sum function $SUM$, state vector $A$, horizontal view of robot $fhfov$, sum of node explorations $EXP$\\
\textbf{Output}: History Score $HS$
\begin{algorithmic}[1] 
\FOR{robot $= 1 \ldots N_{robot}$}
\STATE $min\_dis \gets MAX\_VALUE$
\FOR{$h\subseteq H$}
\IF {$D(x, h) < min\_dis$}
\STATE $min\_dis \gets D(x, h)$
\ENDIF
\ENDFOR
\IF {$min\_dis \geq 25$}
\IF {$A$ is None}
\STATE Create a 360-dimensional zero state vector $A$
\ENDIF
\IF {$hfov \geq \frac{fhfov}{2}$ and $hfov < 360 - \frac{fhfov}{2}$}
\STATE $A[hfov-\frac{fhfov}{2}, hfov+\frac{fhfov}{2}] \gets HFOVS^{t}_{i,hfov}$
\ELSIF{$hfov < \frac{fhfov}{2}$}
\STATE $A[0, hfov+\frac{fhfov}{2}] \gets HFOVS^{t}_{i,hfov}$
\STATE $A[360-hfov-\frac{fhfov}{2}, 360) \gets HFOVS^{t}_{i,hfov}$
\ELSE
\STATE $A[hfov-\frac{fhfov}{2}, 360) \gets HFOVS^{t}_{i,hfov}$
\STATE $A[0, hfov+\frac{fhfov}{2}-360] \gets HFOVS^{t}_{i,hfov}$
\ENDIF
\STATE $HS \gets SUM(A) / EXP$
\ELSE
\STATE Select the closest history node $h_c$ and its corresponding state vector $A$
\STATE \textbf{Go to line 12}
\ENDIF
\ENDFOR
\STATE \textbf{return} $HS$
\end{algorithmic}
\label{alg1}
\end{algorithm}

\section{B Experiment Details}
\subsection{B.1 Hyperparameters}
We show the hyperparameters used for all experiments in Table \ref{table4}.
\begin{table}[ht]
\centering
\begin{tabular}{lc}
\toprule
Parameter &  Value \\
\midrule
VLM Temperature & 1.0 \\
VLM top\_p & 1.0  \\
VLM max\_tokens & 256 \\
long-term navigation goal update interval $\delta$ & 25 \\
Temperature scaling $\tau_{ES}$ & 2.0 \\
Temperature scaling $\tau_{JS}$ & 1.0 \\
horizontal view of robot $fhfov$ & 79 \\
frame width & 640 \\
frame height & 480 \\
\bottomrule
\end{tabular}
\caption{Hyperparameters}
\label{table4}
\end{table}

\subsection{B.2 Computational Resources}
We show the computational resources of the HM3D\_v0.2 experiment in Table \ref{table5}. The MP3D experiment took 23d 03h, and the other settings are consistent. In addition, we recommend readers use higher-performance GPUs  (e.g., RTX 4090) for the experiments, which can effectively reduce the evaluation runtime. \textbf{Since the robots communicate using a global semantic map, we consider the physical time for communication between the robots to be negligible.}
\begin{table}[ht]
\centering
\begin{tabular}{lc}
\toprule
Parameter &  Value \\
\midrule
VLM & GLM-4V-9B~\cite{glm2024chatglm} \\
Quantization Dtype & INT4 \\ 
Evaluation Runtime & 10d 05h \\
Compute Resources & 1 $\times$ RTX 3090 \\
\bottomrule
\end{tabular}
\caption{Parameters and resources required to run a round of two-robot MCoCoNav evaluation on HM3D\_v0.2.}
\label{table5}
\end{table}


\begin{table*}
\centering
\begin{tabular}{lcccc}
\toprule
\multirow{2}{*}{Method} & \multirow{2}{*}{\shortstack{Success Distance}} & \multirow{2}{*}{\shortstack{LLM/VLM}} & \multicolumn{2}{c}{HM3D} \\
\cmidrule{4-5}
&  &  & SPL$\uparrow$ & SR$\uparrow$ \\
\midrule
Random Walking& 0.2 & None & 0.000 & 0.000 \\
Frontier Based~\cite{yamauchi1997frontier} & 0.2 & None & 0.123 & 0.237 \\
Random Samples & 0.2 & None & 0.143 & 0.300 \\
\midrule
SemExp~\cite{chaplot2020object} & 0.2 & None & 0.182 & 0.374 \\
L3MVN~\cite{yu2023l3mvn} & 0.2 & RoBERTa-large (Local) & 0.230 & 0.487 \\
Pixel-Nav~\cite{cai2023bridging} & 1.0 & GPT-4 (Remote) & 0.201 & 0.374\\
ESC~\cite{zhou2023esc} & 0.2 & GPT-3.5 (Remote) & 0.225 & 0.388 \\
Co-NavGPT~\cite{yu2023co} & 0.2 & GPT-3.5 (Remote) & 0.223 & 0.546\\
OpenFMNav~\cite{kuang2024openfmnav} & 0.2 & GPT-4/GPT-4V (Remote) & 0.238 & 0.529 \\
MCoCoNav (Ours) & 0.2 & GLM-4V (Local) & \textbf{0.297} & \textbf{0.634} \\
\bottomrule
\end{tabular}
\caption{\textbf{Comparison of single-robot on different methods.} MCoCoNav significantly outperforms all methods on both benchmarks.}
\label{tab:table6}
\end{table*}

\subsection{B.3 Baseline Details}
\begin{itemize}
\item \textbf{Greedy~\cite{visser2013discussion}:} The Greedy policy assigns frontiers to all robots in a greedy manner. Each robot selects its nearest assigned frontier as its goal position.
\item \textbf{Cost-Utility~\cite{julia2012comparison}:} Each frontier cell $f \in F$ is evaluated using the Cost-Utility method, quantized as $S^{CU}(f)$. We can obtain the $S^{CU}(f)$:
\begin{align}
    S^{CU}(f) &= U(f) - \lambda_{CU}C(f)
\end{align}
where $U(f)$ denotes the utility function that quantifies the size of the frontier area and $C(f)$ denotes the cost function of the distance between the frontier and the robot. The parameter $\lambda_{CU}$ regulates the trade-off between these utility and cost factors. Subsequently, each robot selects the frontier with the highest score of its assigned $S^{CU}$ as its goal location.
\item \textbf{Random Sampling:} This strategy randomly samples each robot's long-term navigation goal on the global semantic map.
\item \textbf{Multi-SemExp~\cite{chaplot2020object}:} The baseline for the multi-robot of ~\cite{chaplot2020object}, where the experiment is set up with two robots collaboratively exploring the environment. The task is considered successful if one of the robots successfully localizes the goal object.
\item \textbf{Co-NavGPT~\cite{yu2023co}:} A typical centralized planning strategy that assigns frontiers to multiple robots for exploration by encoding environmental data as Large Language Model (LLM) prompts.
\item \textbf{GT-Seg:} We replace the semantic segmentation algorithm with ground-truth (GT Seg) to evaluate the relative importance of this module in our framework.
\end{itemize} 

\begin{figure}[t]
    \centering
    \includegraphics[width=8.5cm]{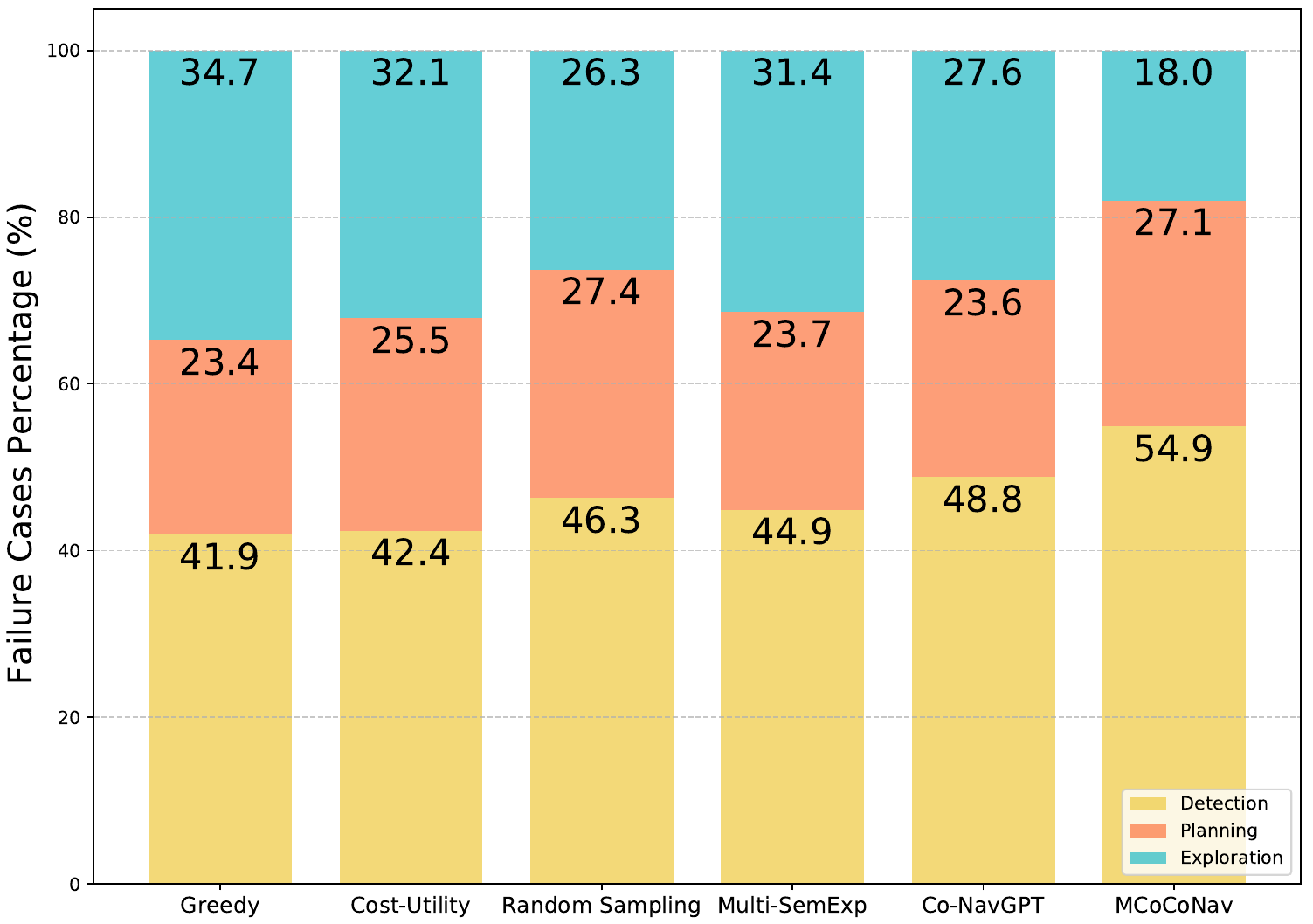}
    \caption{\textbf{Percentage of failure cases for 2-robot in different baselines.} Our MCoCoNav experienced the fewest Exploration failures.}
    \label{fig:2robots}
\end{figure}
\begin{table}
\centering
\begin{tabular}{lccc}
\toprule
\multirow{2}{*}{Method} & \multirow{2}{*}{\shortstack{LLM/VLM}} & \multicolumn{2}{c}{HM3D\_v0.2} \\
\cmidrule{3-4}
&  & SPL$\uparrow$ & SR$\uparrow$ \\
\midrule
Co-NavGPT & GPT-3.5 (Remote) & 0.331 & 0.661\\
Co-NavGPT & GLM-4V (Local) & 0.316 & 0.598\\
MCoCoNav (Ours) & GLM-4V (Local) & \textbf{0.387} & \textbf{0.716} \\
\bottomrule
\end{tabular}
\caption{\textbf{Comparison of different LLM/VLM. } Co-NavGPT as an example.}
\label{tab:table7}
\end{table}
\begin{figure}[t]
    \centering
    \includegraphics[width=8cm]{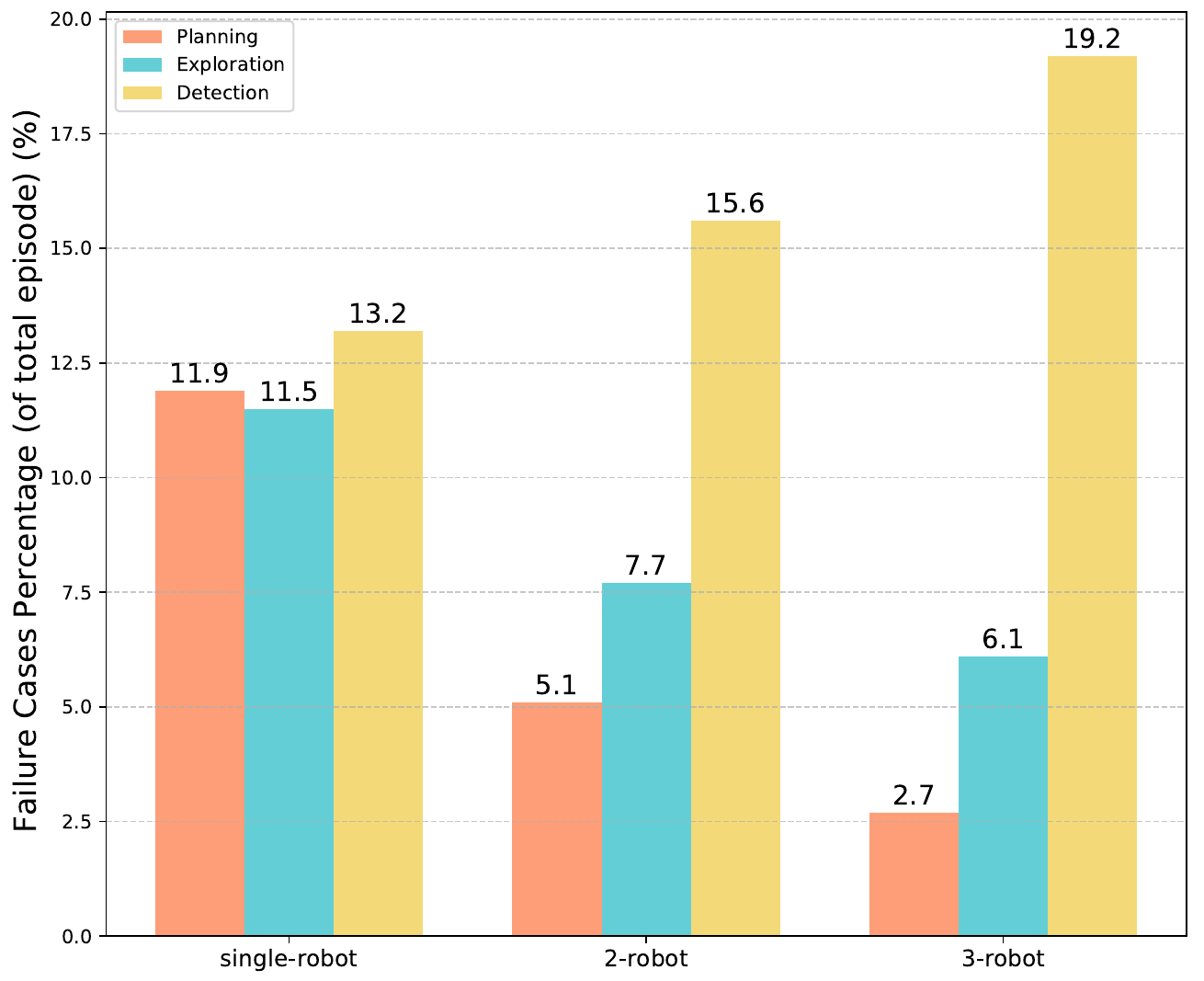}
    \caption{\textbf{Percentage distribution of failure cases in different numbers of robots.} The percentage of detection failures increases with the number of robots.}
    \label{fig:multirobots}
\end{figure}

\subsection{B.4 Comparison of different LLM/VLM}
We conducted experiments on Co-NavGPT~\cite{yu2023co} using the same VLM in table \ref{tab:table7}, and its performance was lower than that of the GPT-3.5 approach. This is due to the low volume and quantized deployment of the GLM-4V. Notably, MCoCoNav demonstrated superior performance even when using a less powerful VLM.

\subsection{B.5 Comparison of Single-Robot}
As shown in Table \ref{tab:table6}, our single-robot approach outperforms the best competitors on HM3D (+3.7\% SPL and +8.5\% in SR). The results indicate that Random Walking fails in nearly all cases without any specialized navigation strategies. In contrast, the significant improvements achieved by SemExp underscore the importance of semantic information in efficient exploration. Furthermore, L3MVN, Co-NavGPT, and ESC use LLMs to select appropriate frontier points, while Pixel-Nav uses LLMs to predict the most likely direction to reach the goal, subsequently employing an RGB-based strategy to plan the route and navigate accordingly. OpenFMNav enhances the navigation process by utilizing a more comprehensive observation of the scene. However, these semantic planning methods do not consider the historical information of the global semantic map and the balance between exploration and return, thus failing to leverage the global semantic information of the environment for planning. Our MCoCoNav fully exploits the VLM's understanding of global semantic information while reconsidering already explored historical nodes, allowing robots to move to locations most likely leading to the goal.

\subsection{B.6 Failure Case Study}
We analyzed all failure instances across episodes and classified them into detection, planning, and exploration failures. Detection failures happen when the robot misidentifies non-goal items as goals or misses actual goals within its view. Planning failures occur when the robot gets stuck or can't reach the goal despite correct detection. Exploration failures happen when the robot doesn't find the goal within the allowed steps. As shown in Figure \ref{fig:2robots}, most navigation failures in the 2-robot setup on HM3D\_v0.2 are due to detection failures, which are caused by inaccurate semantic segmentation or low precision in 3D scanning. In our MCoCoNav, detection failures have the highest occurrence rate. Additionally, as illustrated in Figure \ref{fig:multirobots}, the proportion of detection failures in MCoCoNav increases as the number of robots involved in navigation grows, aligning with our conclusions discussed in the "Effect of the number of robots and semantic accuracy" section.

\subsection{B.7 Prompt Template}
\begin{figure*}[!t]
\begin{center}
  \centerline{\includegraphics[width=18cm]{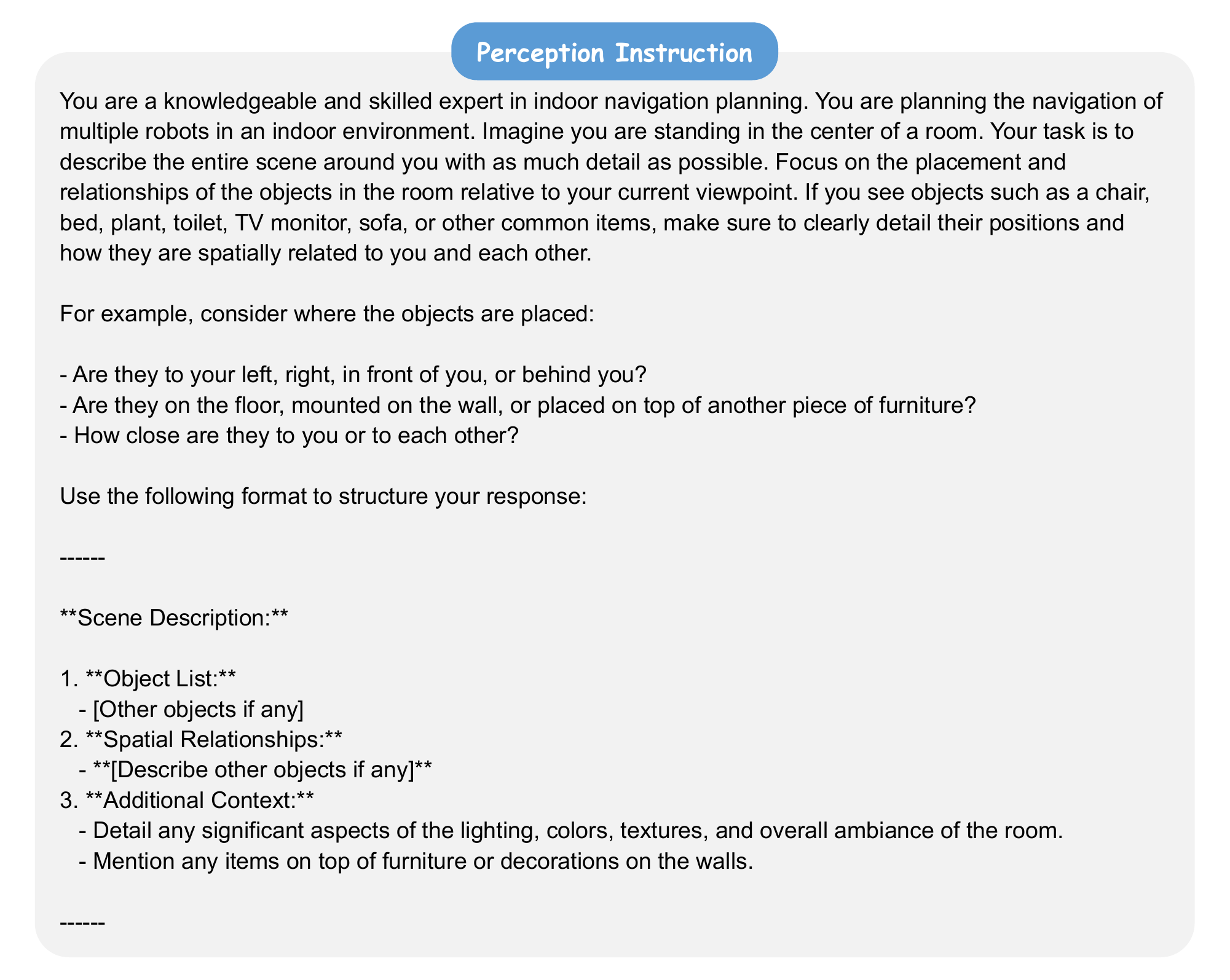}}
\end{center}
	\caption{Perception Instruction Template.}
\label{fig:exp-4}
\end{figure*}

\begin{figure*}[!t]
\begin{center}
  \centerline{\includegraphics[width=18cm]{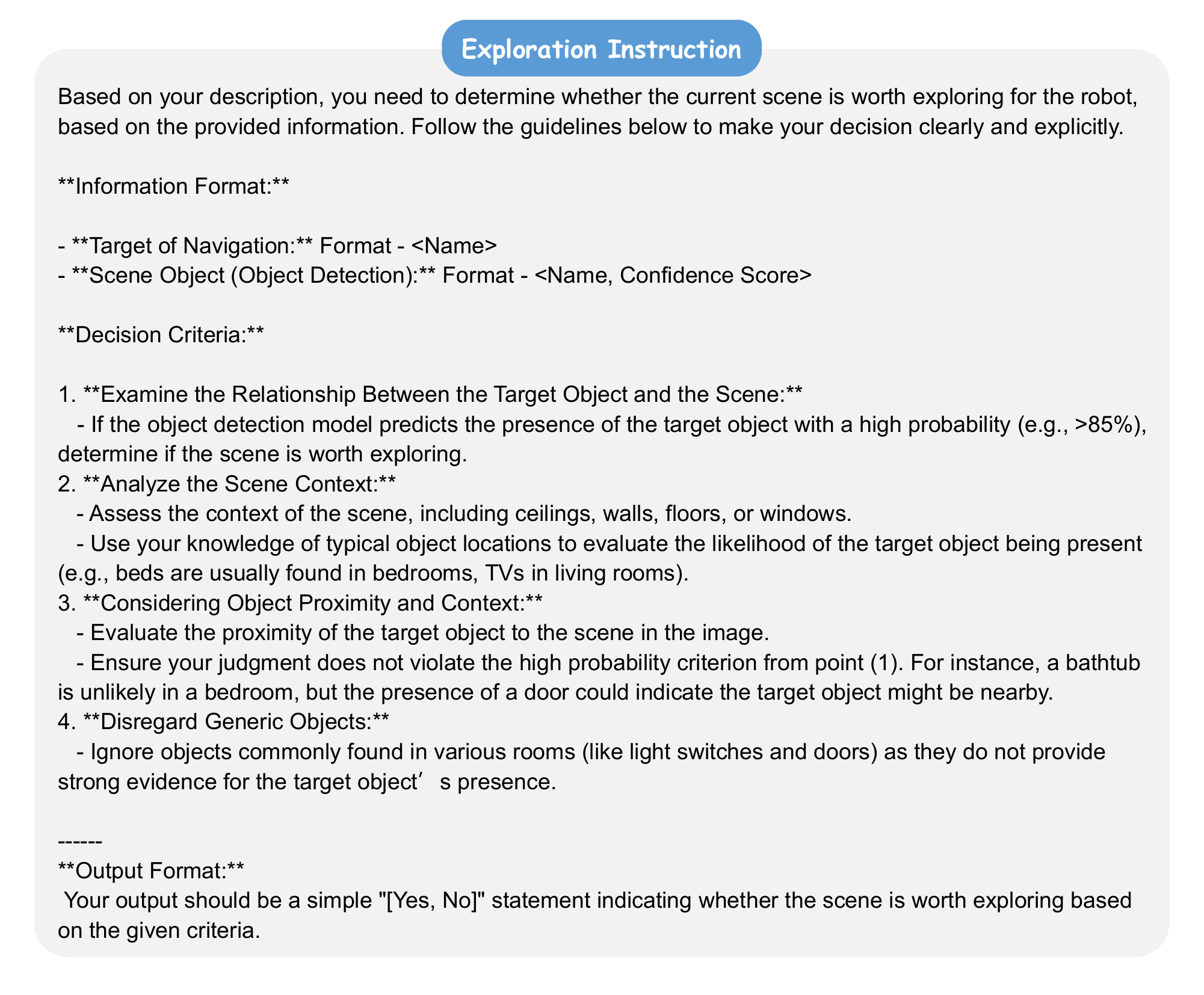}}
\end{center}
	\caption{Exploration Instruction Template.}
\label{fig:exp-5}
\end{figure*}

\begin{figure*}[!t]
\vspace{-11pt}
\begin{center}
  \centerline{\includegraphics[width=18cm]{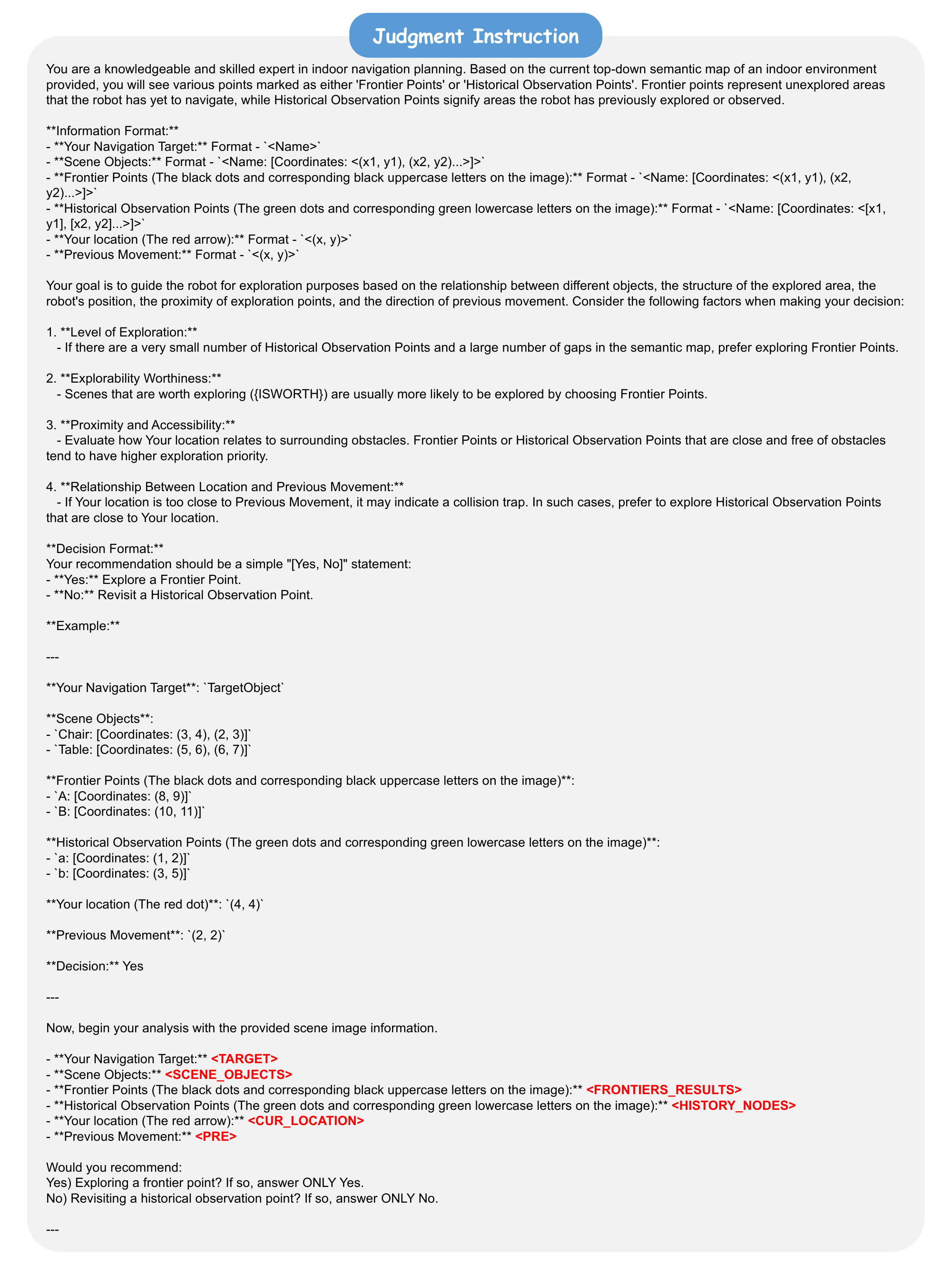}}
\end{center}
	\caption{Judgment Instruction Template.}
\label{fig:exp-6}
\end{figure*}

\begin{figure*}[!t]
\begin{center}
  \centerline{\includegraphics[width=18cm]{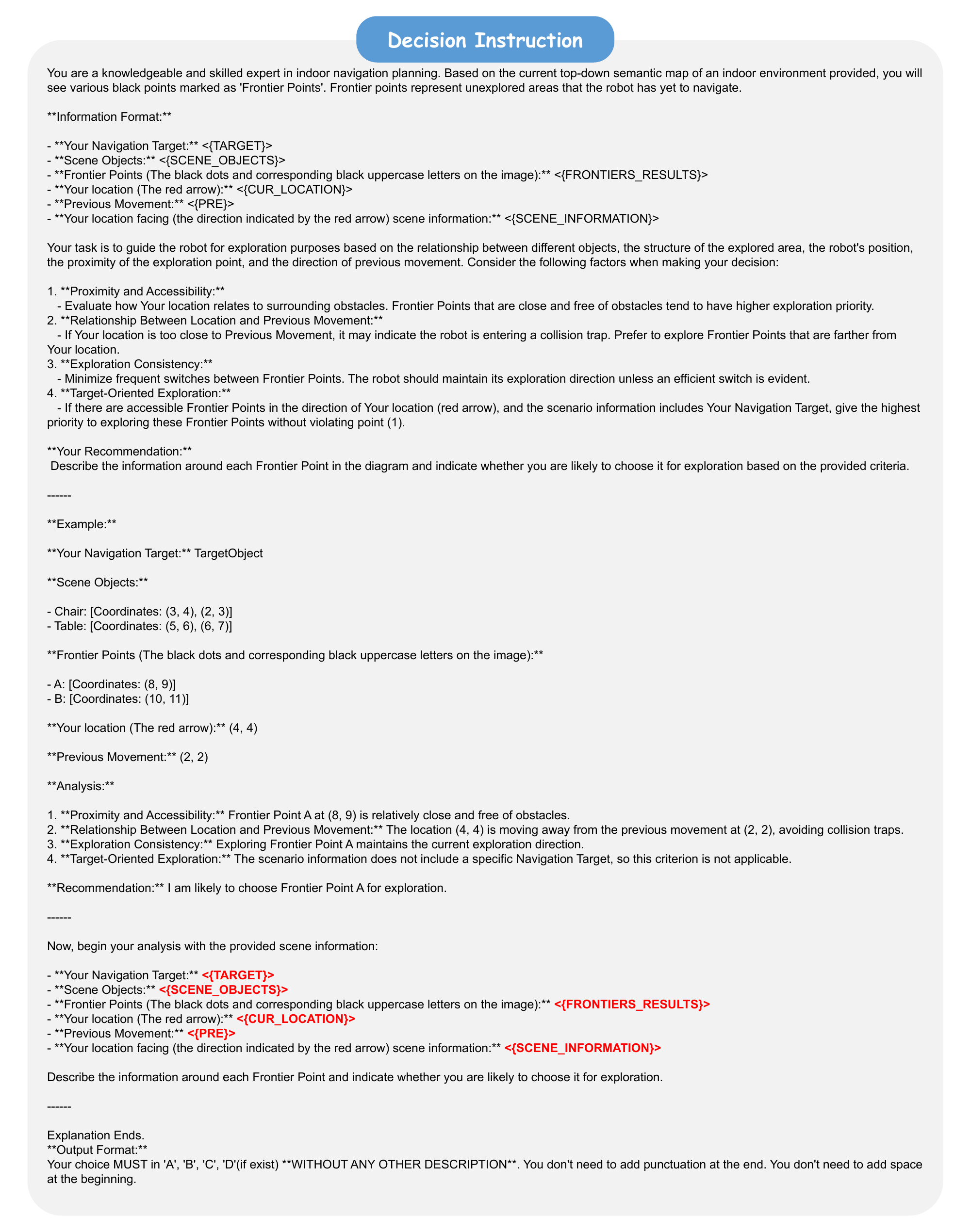}}
\end{center}
	\caption{Decision Instruction Template.}
\label{fig:exp-7}
\end{figure*}

From Figure \ref{fig:exp-4} to \ref{fig:exp-7}, we give all the specific prompt templates for the cross-image multimodal CoT. The red fonts in the prompt templates represent different segments in different scenarios.

\end{document}